\documentclass{article}

\usepackage[preprint]{neurips_2025}

\usepackage[utf8]{inputenc} 
\usepackage[T1]{fontenc}    
\usepackage{hyperref}       
\usepackage{url}            
\usepackage{booktabs}       
\usepackage{amsfonts}       
\usepackage{nicefrac}       
\usepackage{microtype}      
\usepackage{xcolor}         
\usepackage{graphicx}
\usepackage{subcaption}  
\usepackage{caption}     
\usepackage{amsmath}
\usepackage{multirow}
\usepackage{booktabs}
\usepackage{pifont}

\title{OCRVerse: Towards Holistic OCR in End-to-End  Vision-Language Models}

\author{
\textbf{Yufeng Zhong}\textsuperscript{\textbf{1}}\thanks{Equal Contribution.}, 
\textbf{Lei Chen}\textsuperscript{\textbf{1}}\footnotemark[1], 
\textbf{Xuanle Zhao}\textsuperscript{\textbf{1}}\footnotemark[1], 
\textbf{Wenkang Han}\textsuperscript{\textbf{1}}, 
\textbf{Liming Zheng}\textsuperscript{\textbf{1}}, \\
\textbf{Jing Huang}\textsuperscript{\textbf{1}},
\textbf{Deyang Jiang}\textsuperscript{\textbf{1}}, 
\textbf{Yilin Cao}\textsuperscript{\textbf{1}},
\textbf{Lin Ma}\textsuperscript{\textbf{1}}\thanks{Corresponding Author.}, 
\textbf{Zhixiong Zeng}\textsuperscript{\textbf{1}}\footnotemark[2] \\
\textsuperscript{1}Meituan,\\
\texttt{forest.linma@gmail.com, zengzhixiong@meituan.com}
}

\begin{document}

\maketitle
\begin{figure}[h]
    \centering
    \vspace{-10pt}
    \begin{subfigure}[b]{0.9\textwidth}
        \centering
        \includegraphics[width=\textwidth]{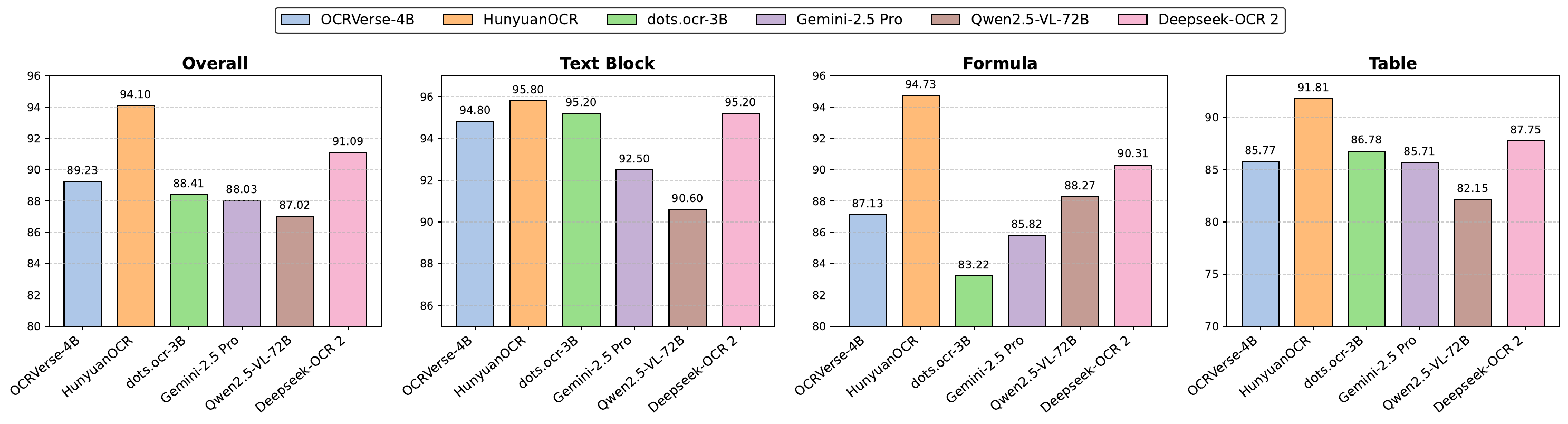}
        \label{fig:text_bar}
    \end{subfigure}
    \begin{subfigure}[b]{0.9\textwidth}
        \centering
        \includegraphics[width=\textwidth]{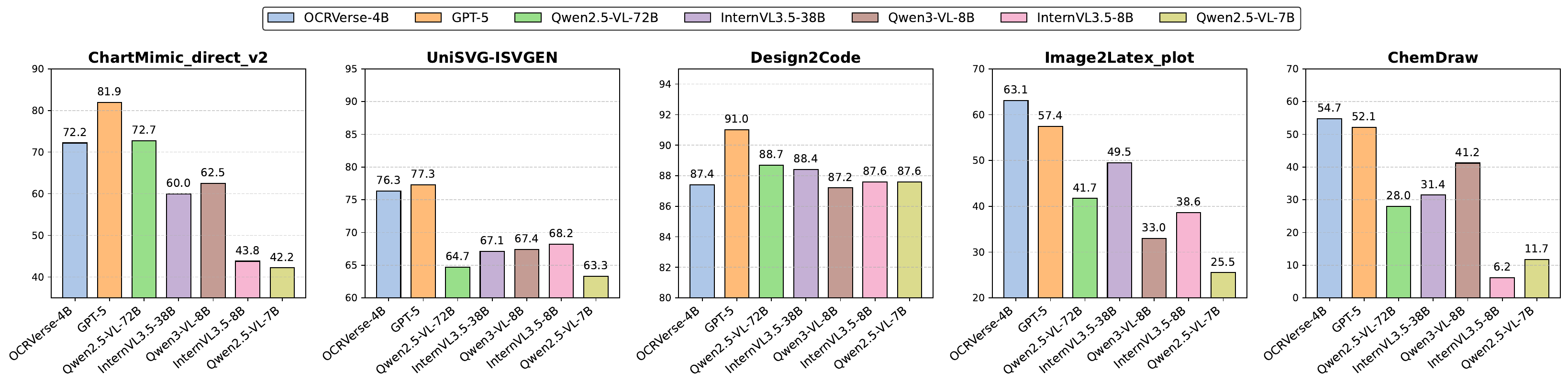}
        \label{fig:code_bar}
    \end{subfigure}
    \vspace{-10pt}
    \caption{
    Performance comparison of OCRVerse on text-centric OCR tasks (top row) and vision-centric OCR tasks (bottom row). Since existing OCR methods primarily focus on text-centric scenarios, we compare against both specialized OCR models and general-purpose models for text-centric benchmarks, while comparing only against general-purpose models for vision-centric benchmarks.
    }
    \label{fig:first_bar}
\end{figure}

\begin{abstract}
The development of large vision language models drives the demand for managing, and applying massive amounts of multimodal data, making OCR technology, which extracts information from visual images, increasingly popular. However, existing OCR methods primarily focus on recognizing text elements from images or scanned documents (\textbf{Text-centric OCR}), neglecting the identification of visual elements from visually information-dense image sources (\textbf{Vision-centric OCR}), such as charts, web pages and science plots. In reality, these visually information-dense images are widespread on the internet and have significant real-world application value, such as data visualization and web page analysis.
In this technical report, we propose \textbf{OCRVerse}, the first holistic OCR method in end-to-end manner that enables unified text-centric OCR and vision-centric OCR.
To this end, we construct comprehensive data engineering to cover a wide range of text-centric documents, such as newspapers, magazines and books, as well as vision-centric rendered composites, including charts, web pages and scientific plots.
Moreover, we propose a two-stage SFT-RL multi-domain training method for OCRVerse. SFT directly mixes cross-domain data to train and establish initial domain knowledge, while RL focuses on designing personalized reward strategies for the characteristics of each domain. Specifically, since different domains require various output formats and expected outputs, we provide sufficient flexibility in the RL stage to customize flexible reward signals for each domain, thereby improving cross-domain fusion and avoiding data conflicts.
Experimental results demonstrate the effectiveness of OCRVerse, achieving competitive results across text-centric and vision-centric data types, even comparable to large-scale open-source and closed-source models.

\end{abstract}

\section{Introduction}

The rapid development of Large Vision Language Models (LVLM)~\cite{bai2025qwen25vl,zhu2025internvl3,anthropic2025claude,openai2025gpt5,comanici2025gemini}  has driven unprecedented growth in applications that manage and apply massive amounts of multimodal data. These applications increasingly demand the ability to extract and process textual information from diverse visual sources, making Optical Character Recognition (OCR) technology more critical than ever. As a fundamental bridge connecting visual content to language models, OCR enables a wide range of applications including document digitization~\cite{sinha2025digitization}, automated data entry~\cite{amgad2025digitization}, and intelligent content analysis~\cite{huang2023intelligent}.

Initially, OCR technology primarily focused on recognizing text elements from images or scanned documents, which we term \textbf{Text-centric OCR}. Current text-centric OCR methods can be categorized into two major types, namely pipeline-based methods~\cite{paruchuri2025marker,cui2025paddleocr3,wang2024mineru,wang2021pgnet,zhang2018multi} and Vision-Language Model (VLM)-based methods~\cite{li2025dots,blecher2023nougat,hu2025mplug,liu2024focus,wei2024vary}. Pipeline-based methods utilize expert modules in a two-stage process where the layout analysis module first identifies key regions and then specialized parser extracts content from each sub-region. While offering high stability, this approach suffers from limited flexibility and high fine-tuning costs. Conversely, VLM-based methods employ end-to-end architectures that decode text directly from visual inputs, thereby providing simplified workflows and superior generalization. However, lacking explicit layout modeling, VLM-based methods often produce hallucinations such as incorrect reading orders and missing content. Recent hybrid approaches~\cite{niu2025mineru2,cui2025paddleocrvl,chen2025logics,feng2025dolphin,li2025monkeyocr} combine traditional OCR for layout analysis with VLMs for content understanding, thus achieving superior performance on documents containing complex elements such as tables and formulas.

Although text-centric OCR has achieved significant breakthroughs in document processing, its application scope remains limited to traditional document types. However, as industries digitalize and the internet proliferates with diverse visual content, existing OCR methods neglect the identification of visual elements from visually information-dense image sources such as charts~\cite{yang2024chartmimic, chen2025breaking}, web pages~\cite{si2025design2code} and scientific plots~\cite{zhao2025vincicoder}, which we define as \textbf{Vision-centric OCR}. Unlike traditional documents that primarily contain plain text, vision-centric scenarios exhibit unique dual characteristics where they contain both conventional printed text and rich visual elements such as arrows, lines, and icons that build semantic structures expressing logical relationships beyond plain text. Consequently, traditional text-centric OCR becomes inadequate because these scenarios require capturing not only character but also semantic relationships embedded in visual structures. To achieve this, code-level representations become essential where HTML code encodes webpage layouts, Python code represents computational logic in charts, and LaTeX captures mathematical semantics in scientific plots~\cite{zhong2025doctron}. In reality, these visually information-dense images are widespread on the internet and have significant real-world application value in data visualization and web page analysis.

To address the limitations of fragmented approaches, holistic OCR has emerged as a paradigm that unifies both text-centric and vision-centric capabilities, integrating character-level and code-level representations. However, traditional pipeline-based methods face inherent limitations in understanding professional visual semantics, making them inadequate for such complex scenarios. In contrast, VLM-based methods pre-trained on large-scale visual data leverage powerful cross-modal semantic understanding, thereby offering new solutions for advancing OCR technology toward holistic recognition. 

To this end, we propose OCRVerse, the first holistic OCR method in an end-to-end manner that enables unified text-centric OCR and vision-centric OCR. Specifically, we build OCRVerse through two complementary strategies, namely comprehensive data engineering and an innovative two-stage SFT-RL multi-domain training method. On the data front, we construct comprehensive data engineering to cover a wide range of text-centric documents such as newspapers, magazines and books, as well as vision-centric rendered composites including charts, web pages, and scientific plots. On the model front, we develop a lightweight OCR model based on Qwen3-VL 4B~\cite{Qwen3-VL} and propose a two-stage SFT-RL multi-domain training method. In the SFT stage, we aim to establish foundational cross-domain knowledge by directly mixing data from all domains. This approach enables the model to learn diverse visual patterns and output formats across both text-centric and vision-centric scenarios, thereby building a unified representation space. In the RL stage, we aim to resolve domain-specific conflicts and optimize personalized performance for each domain. Since different domains require various output formats and quality expectations, we design customized reward strategies tailored to the characteristics of each domain, such as structural accuracy for tables and semantic fidelity for charts. This flexible reward mechanism effectively improves cross-domain fusion while avoiding data conflicts that typically arise from naive multi-task learning. As shown in Figure~\ref{fig:first_bar}, experimental results demonstrate the effectiveness of OCRVerse, achieving competitive results across both text-centric and vision-centric domains, even comparable to large-scale open-source and closed-source models.

The main contributions of this study are as follows:

\begin{itemize}

\item \textbf{Holistic OCR Paradigm:} We propose OCRVerse, the first end-to-end holistic OCR method that unifies text-centric and vision-centric capabilities, bridging character-level recognition and code-level representation through a lightweight architecture.

\item \textbf{Two-Stage SFT-RL Multi-Domain Training:} We introduce an innovative two-stage SFT-RL training methodology where SFT establishes foundational cross-domain knowledge while RL employs personalized reward strategies to resolve domain conflicts, effectively enabling seamless fusion across eight diverse data types.

\item \textbf{Strong Empirical Performance:} Experimental results demonstrate competitive performance across both text-centric and vision-centric scenarios, achieving 89.23 on OmniDocBench v1.5~\cite{ouyang2025omnidocbench} and comparable results to open-source models across multiple vision-centric benchmarks.

\end{itemize}

\section{Related Work}
\subsection{Text-centric OCR}
Text-centric OCR aims to extract and convert structured textual information from document images, serving as a pivotal technology for digitizing vast human knowledge and supplying high-quality corpora for large language models (LLMs). From the perspective of architectural evolution, current methodologies in this field can be systematically categorized into three paradigms: (1) Traditional pipeline methods~\cite{paruchuri2025marker,cui2025paddleocr3,wang2024mineru,wang2021pgnet,zhang2018multi}, which decompose the parsing task into cascaded specialized modules for  layout detection and text recognition; (2) VLM-based end-to-end methods~\cite{li2025dots,blecher2023nougat,hu2025mplug,liu2024focus,wei2024vary}, which leverage unified Transformer architectures~\cite{vaswani2017attention} to directly map visual inputs into serialized text sequences; and (3) VLM-based pipeline methods~\cite{niu2025mineru2,cui2025paddleocrvl,chen2025logics,feng2025dolphin,li2025monkeyocr}, designed to synergize explicit layout priors with the semantic reasoning capabilities of VLMs.

Traditional pipeline approaches adopt a ``divide-and-conquer'' strategy, utilizing specialized modules for layout detection and text recognition. Representative tools like Marker~\cite{paruchuri2025marker} and PP-StructureV3~\cite{cui2025paddleocr3} employ heuristic rules or lightweight detectors to localize regions (\textit{e.g.}, tables, formulas) before applying dedicated OCR engines. MinerU (Pipeline)~\cite{wang2024mineru} further refines this process with high-precision PDF parsing engineering to ensure structural consistency. Despite their stability and interpretability, these methods suffer from cascading errors—where layout detection failures propagate to recognition—and limited generalization on complex, unconstrained documents due to their reliance on rigid engineering heuristics.

VLM-based end-to-end methods simplify the workflow by leveraging the generative capabilities of Transformers to decode text directly from visual features. Nougat~\cite{blecher2023nougat} pioneered this paradigm for academic documents parsing, employing a Swin Transformer~\cite{liu2021swin} encoder and BART~\cite{lewis2020bart} decoder. More recently, DeepSeek-OCR~\cite{wei2025deepseekocr} and GOT-OCR~\cite{wei2024general} have scaled this approach, achieving impressive performance on diverse document types through visual compression and unified training. While these models exhibit strong semantic reasoning, they face challenges with high-entropy content (\textit{e.g.}, dense tables and complex formulas), often resulting in hallucinations, repetition, or attention drift during long-sequence generation.

VLM-based pipeline methods emerge as a hybrid solution, integrating explicit layout priors to guide the semantic reasoning of LVLMs. MinerU 2.5~\cite{niu2025mineru2} and PaddleOCR-VL~\cite{cui2025paddleocrvl} utilize detectors to crop and order document regions, allowing the VLM to process high-resolution inputs with reduced noise. Similarly, MonkeyOCR~\cite{li2025monkeyocr} employs a structure-aware relation parser to enhance the model's understanding of complex layouts. Although these hybrid frameworks effectively mitigate hallucinations and resolution constraints, they primarily remain text-centric, often neglecting the holistic interpretation of vision-intensive elements such as scientific plots and web UI components, which is a core focus of our work.

\vspace{-2mm}

\subsection{Vision-centric OCR}

\vspace{-1mm}



Diverging from the character-level focus of text-centric OCR, vision-centric OCR 
targets the interpretation of information-dense sources such as charts, webpages, scientific plots, and scalable vector graphics. By translating visual pixels into executable code—including HTML, LaTeX, and Python—this paradigm yields structured representations that either support rapid re-creation or serve as valuable corpora for deep learning.

Website and Graphical User Interface Parsing. In the realm of website and graphical user interface parsing, the primary objective is to transform intricate visual hierarchies into structured markup languages. Early endeavors like Pix2Code~\cite{beltramelli2018pix2code} and Sketch2Code~\cite{jain2019sketch2code} pioneered the mapping of graphical screenshots to domain-specific languages by utilizing modular perception units. Driven by the reasoning capabilities of multimodal LLMs, recent research has transitioned toward direct visual-to-code translation, such as WebSight~\cite{laurenccon2024unlocking} and Design2Code~\cite{si2025design2code}, emphasizing the reconstruction of visual fidelity through large-scale supervised fine-tuning on synthetic and real-world web corpora. To overcome the resolution bottlenecks and structural hallucinations inherent in single-pass models, hybrid frameworks like CogAgent~\cite{hong2024cogagent} and EfficientUICoder~\cite{xiao2025efficientuicoder} introduce high-resolution encoders and token compression strategies to better discern minute user interface components and complex spatial arrangements.


Scientific Visualization. Translating charts and scientific illustrations into executable code transforms data science by converting static visual assets into editable, programmatic representations that enhance knowledge reuse. In chart interpretation, benchmarks such as Plot2Code~\cite{wu2025plot2code} and ChartMimic~\cite{yang2024chartmimic} provide rigorous standards for mapping complex visual hierarchies to ground-truth code. Complementary generative frameworks, notably ChartMaster~\cite{tan2025chartmaster}, employ large-scale supervised training and reinforcement learning with multimodal feedback to achieve syntactic and structural alignment. This paradigm extends to specialized scientific demonstrations, such as ChemDraw~\cite{zhao2025vincicoder}, which translates molecular diagrams into executable visualization scripts. 

Scalable Vector Graphics Generation. Scalable vector graphics (SVG) generation provides resolution-independent and editable representations that link visual concepts with structured primitives. This field has evolved from inverse graphics optimization using differentiable rasterizers~\cite{li2020differentiable, ma2022towards} and sequence modeling~\cite{wu2023iconshop} toward structural reasoning with vision-language models. Foundational frameworks such as StarVector\cite{rodriguez2025starvector} and OmniSVG~\cite{yang2025omnisvg} unify multimodal generation through large-scale synthesis and primitive-aware parameterization. To improve structural fidelity, recent works like Reason-SVG~\cite{xing2025reason} and RLRF~\cite{rodriguez2025rendering} incorporate reinforcement learning and rendering-aware optimization to refine the generated code for high-fidelity visual reconstruction.

Despite these advancements, existing methods predominantly operate in silos, constrained to specific tasks and singular output formats. They lack a unified framework capable of handling the diverse spectrum of vision-centric OCR tasks simultaneously, failing to achieve cross-scenario generalization and unified processing of multi-type visual information-dense images. This fragmentation motivates OCRVerse, a holistic framework that unifies text-centric and vision-centric OCR.



\section{Dataset}

\subsection{Data Types}
\begin{figure*}[!t]
\centering
\includegraphics[width=\textwidth]{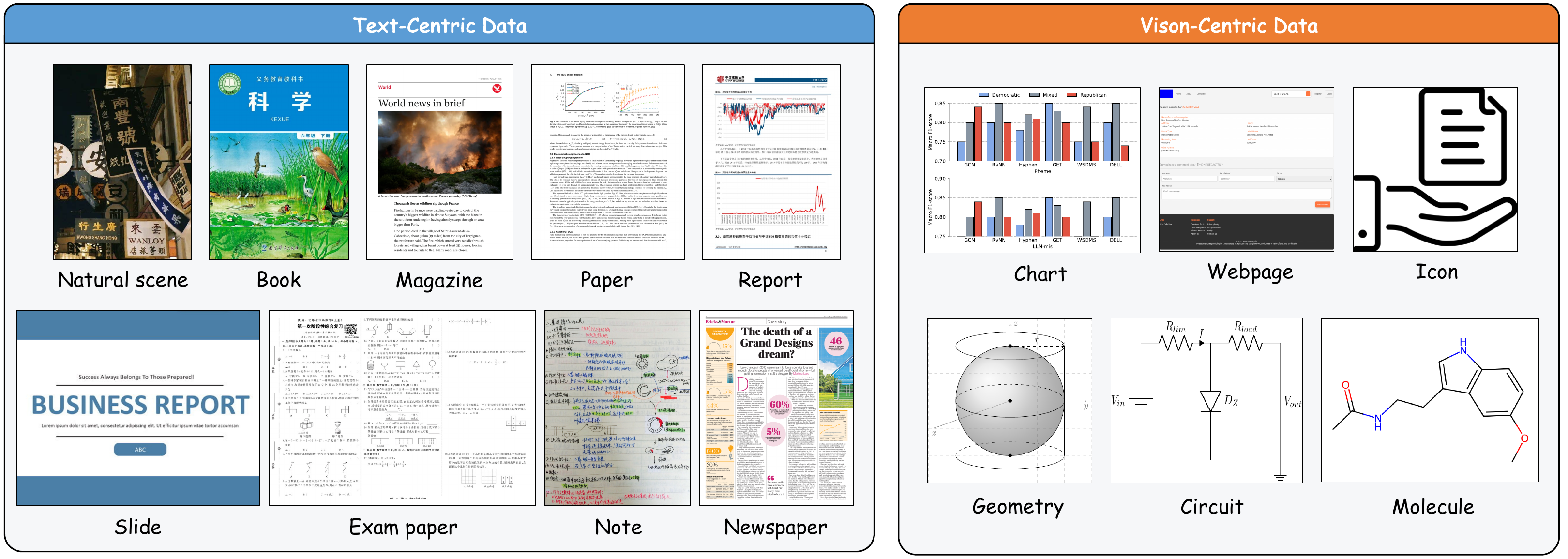}
\caption{Comprehensive data coverage of OCRVerse for holistic OCR. \textbf{Left (Text-centric data):} Nine document scenarios including natural scenes, books, magazines, papers, reports, slides, exam papers, notes, and newspapers, covering high-frequency text scenarios in daily life. \textbf{Right (Vision-centric data):} Six specialized scenarios including charts, webpages, icons, geometry, circuits, and molecules, focusing on professional structured content that requires code-level representations.}
\vspace{-2mm}
\label{fig:data_source}
\end{figure*}

As illustrated in Figure~\ref{fig:data_source}, OCRVerse encompasses both text-centric and vision-centric data types, comprehensively supporting the data requirements of holistic OCR. The text-centric data types cover nine document scenarios: natural scenes, books, magazines, papers, reports, slides, exam papers, notes, and newspapers, which encompass high-frequency text scenarios in daily life and meet essential OCR needs. The vision-centric data types comprise six specialized scenarios: charts, webpages, icons, geometry, circuits, and molecules, which focus on professional structured content and address gaps not covered by text-centric categories.

\subsubsection{Text-centric Data Types}

Our text-centric dataset encompasses nine distinct document domains, each exhibiting unique characteristics in layout organization, content structure, and information density. \textbf{Natural scenes} capture text that appears in real-world environments such as street signs, product packaging, and storefront displays, which feature varied fonts, orientations, and complex backgrounds. \textbf{Books} present continuous narrative text that is organized in chapters, along with footnotes and occasional illustrations. \textbf{Magazines} integrate stylized typography with rich visual elements, which blend editorial content with advertisements in creative layouts. \textbf{Papers} typically follow standardized academic formats that include multi-column arrangements, mathematical expressions, citations, and integrated figures and tables. \textbf{Reports} deliver structured information with emphasis on data tables and statistical charts, which are organized in formal layouts for professional communication. \textbf{Slides} contain concise bullet points, large fonts, and visual aids that are designed for presentation contexts. \textbf{Exam papers} contain diverse question formats including multiple-choice, fill-in-the-blank, and problem-solving items, which often include complex mathematical notations. \textbf{Notes} exhibit informal handwriting or typed annotations with non-standard layouts and personal organizational styles.This comprehensive coverage across document types captures the structural and semantic diversity that exists in real-world text scenarios, which enables reliable parsing of diverse layouts and effective information extraction across different formats.


\subsubsection{Vison-centric Data Types}Our vision-centric dataset targets six specialized domains that require structured representation and semantic understanding. \textbf{Charts} encompass various visualization types including bar charts, line graphs, pie charts, and scatter plots, requiring both visual element recognition and understanding of data relationships. \textbf{Webpages} preserve hierarchical structures when converted to images, and they integrate navigation menus, hyperlinks, forms, and dynamic content layouts that characterize digital interfaces. \textbf{Icons} are symbolic graphics that appear in interfaces and signage systems, where recognition of abstract shapes and their symbolic meanings is essential. \textbf{Geometry} includes mathematical diagrams, geometric constructions, and spatial relationships, all of which necessitate precise shape recognition and spatial reasoning capabilities. \textbf{Circuits} consist of electronic diagrams using standardized symbols, connection lines, and component labels, demanding knowledge of electrical engineering notation systems. \textbf{Molecules} display chemical structures with atomic bonds, functional groups, and stereochemical notations, which require understanding of chemistry-specific visual languages.These professional domains extend beyond text-centric categories by capturing structured and symbolic content that traditional OCR cannot handle, thereby providing essential supplements for comprehensive holistic OCR capabilities.

\subsection{Data Processing}
\begin{figure*}[!t]
\centering
\includegraphics[width=\textwidth]{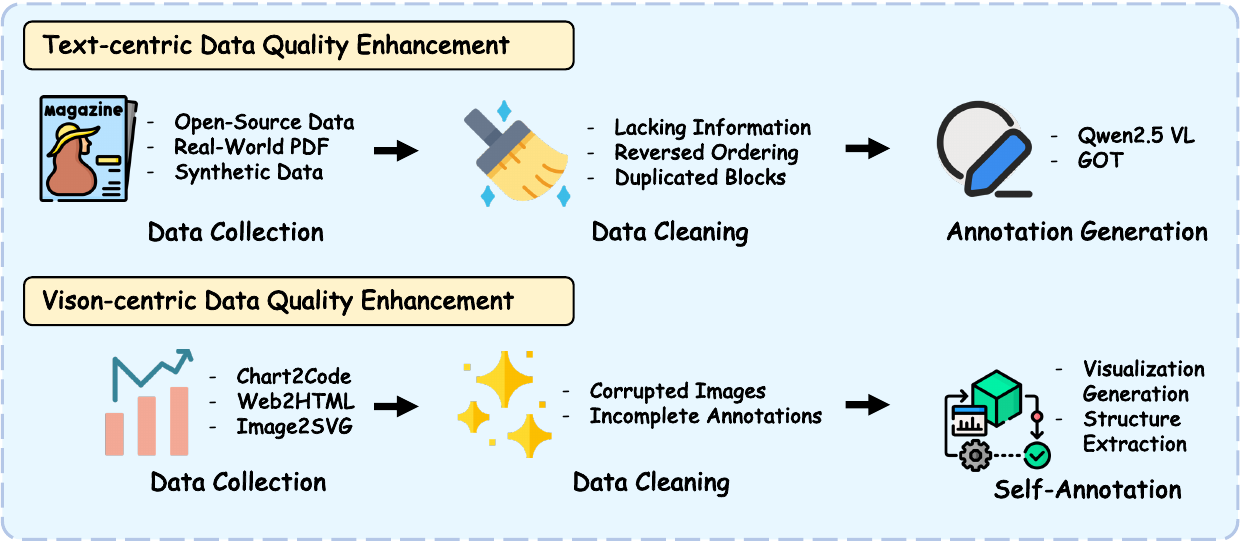}
\caption{Multi-stage data construction pipeline integrating text-centric and vision-centric sources. Text-centric pipeline (top) processes open-source data, real-world PDFs, and synthetic data through cleaning and VLM-based annotation. Vision-centric pipeline (bottom) collects chart, webpage, and SVG data, etc., applies quality filtering, and generates annotations through visualization rendering and structure extraction.}

\vspace{-5mm}
\label{fig:data_pipeline}
\end{figure*}

As illustrated in Figure~\ref{fig:data_pipeline}, our training dataset is constructed through a systematic multi-stage pipeline that integrates both text-centric and vision-centric data sources to ensure comprehensive coverage and high quality.

\subsubsection{Text-centric Data Quality Enhancement}

To construct high-quality text-centric OCR data, we integrate three data sources through systematic processing pipelines: open-source datasets for large-scale coverage, real-world PDF documents for authentic scenario alignment, and synthetically generated data for domain-specific augmentation.

\noindent \textbf{Data Collection.} We collect diverse text-centric datasets spanning multiple domains to support comprehensive document understanding. Our open-source data encompasses natural scene text from LSVT~\cite{sun2019icdar} and TextOCR~\cite{singh2021textocr}, document data from PDFA~\cite{pdfa}, DocStruct4M~\cite{hu2024mplug}, and DocGenome~\cite{xia2024docgenome}, as well as handwritten text from IAM~\cite{marti2002iam}, ORAND-CAR~\cite{diem2014icfhr}, and HME~\cite{yuan2022syntax} collections. Real-world PDF documents are collected from books, magazines, academic papers, reports, and presentation slides, providing authentic layout structures and content patterns. Synthetic data is constructed from subject examination questions across K12 to graduate levels and mathematical formula question-answer pairs from StackExchange, enabling targeted augmentation for specialized content types.

\noindent \textbf{Data Cleaning.} To ensure consistent data quality across heterogeneous sources, we implement systematic cleaning strategies tailored to different data characteristics. We first perform general quality inspection to identify and remove samples with missing content, incorrect reading order, and repeated paragraphs that compromise annotation reliability. For document data, we split multi-page PDFs into individual pages and categorize them based on task requirements, distinguishing between layout analysis tasks that focus on structure understanding and content parsing tasks that emphasize text recognition. For examination and formula data, we extract structured components through regex patterns combined with manual verification, filter samples containing embedded images, and categorize formulas by complexity levels to enable progressive learning. Through these targeted cleaning procedures, we establish uniform quality standards across all data sources.

\noindent \textbf{Annotation Generation.} To generate high-quality annotations at scale, we adopt diverse annotation strategies matched to data characteristics and task requirements. For existing datasets with quality issues, we employ VLM-based re-annotation using advanced models such as Qwen2.5-VL-72B~\cite{Qwen2.5-VL} and GOT~\cite{wei2024general} to correct errors and enhance annotation completeness. For real-world documents, we utilize specialized OCR tools to extract structured content, generating page-level layouts with bounding boxes, region-level parsing with color-guided localization, and multi-page sampling for cross-page understanding, while converting formulas to LaTeX and tables to HTML for structured representation. For synthetic data, we design parameterized HTML templates that control comprehensive visual attributes, inject content with MathJax rendering for mathematical expressions and CSS styling for structured elements, then perform headless browser rendering to generate image-annotation pairs with Markdown format. Through these complementary annotation approaches, we construct large-scale training data covering diverse document types and complexity levels.

\subsubsection{Vision-centric Data Quality Enhancement}

To ensure high-quality training data for vision-centric content understanding, we systematically enhance open-source vision-centric datasets through a unified processing pipeline spanning data collection, cleaning and self-annotation, with domain-specific augmentation to address coverage gaps.

\noindent \textbf{Data Collection.} We collect diverse vision-centric datasets spanning five specialized domains to support structured content understanding. Our chart-to-code data is sourced from MCD~\cite{jiang2025viscodex} and MSRL~\cite{chen2025breaking}, providing diverse chart types for code generation training. For webpage structure extraction, we integrate datasets from MCD~\cite{jiang2025viscodex}, Web2M~\cite{gui2025webcode2m}, and Web2Code~\cite{yun2024web2code}, which offer comprehensive HTML structure annotations. The vector graphics domain utilizes the ISVGEN subset from UniSVG~\cite{li2025unisvg} as the primary training corpus for SVG generation. Mathematical diagram data is curated from DaTikZ-v3~\cite{belouadi2024detikzify} and Cosyn-400k~\cite{yang2025scaling}, covering geometric constructions and TikZ-based representations. For molecular structure understanding, we construct our dataset by combining samples from the Cosyn-400k~\cite{yang2025scaling} collection with various open-source text-to-mermaid datasets, enabling chemical notation generation.

\noindent \textbf{Data Cleaning.} To ensure high-quality training data for code generation, we implement systematic cleaning strategies that remove low-quality samples across all domains. We first perform general filtering to eliminate corrupted images and incomplete annotations that hinder reliable code generation. Beyond these general procedures, we apply specialized cleaning for specific domains. For webpages, we remove embedded images from HTML annotations, thereby enabling the model to focus on structural understanding rather than visual content. For mathematical diagrams, we enhance the dataset by adding complete LaTeX/TikZ environment rendering capabilities with necessary packages and declarations. Through these comprehensive cleaning procedures, we establish consistent data quality across all domains, providing a solid foundation for subsequent annotation and training.

\noindent \textbf{Self-annotation.} To scale annotation coverage beyond manually labeled data, we adopt a bootstrapping approach that leverages domain-specific models for automated annotation generation. We first train specialized models on cleaned subsets to establish baseline capabilities for each domain, including chart-to-code models for visualization generation, webpage-to-HTML models for structure extraction, image-to-SVG models for vector graphics, and image-to-LaTeX models for mathematical diagrams. These trained models then perform self-annotation on remaining unlabeled samples, generating corresponding code representations that match domain-specific requirements. Through this bootstrapping strategy, we significantly expand training data coverage while maintaining annotation quality consistent with manually curated samples.

\section{Method}

\begin{figure*}[!t]
\centering
\includegraphics[width=\textwidth]{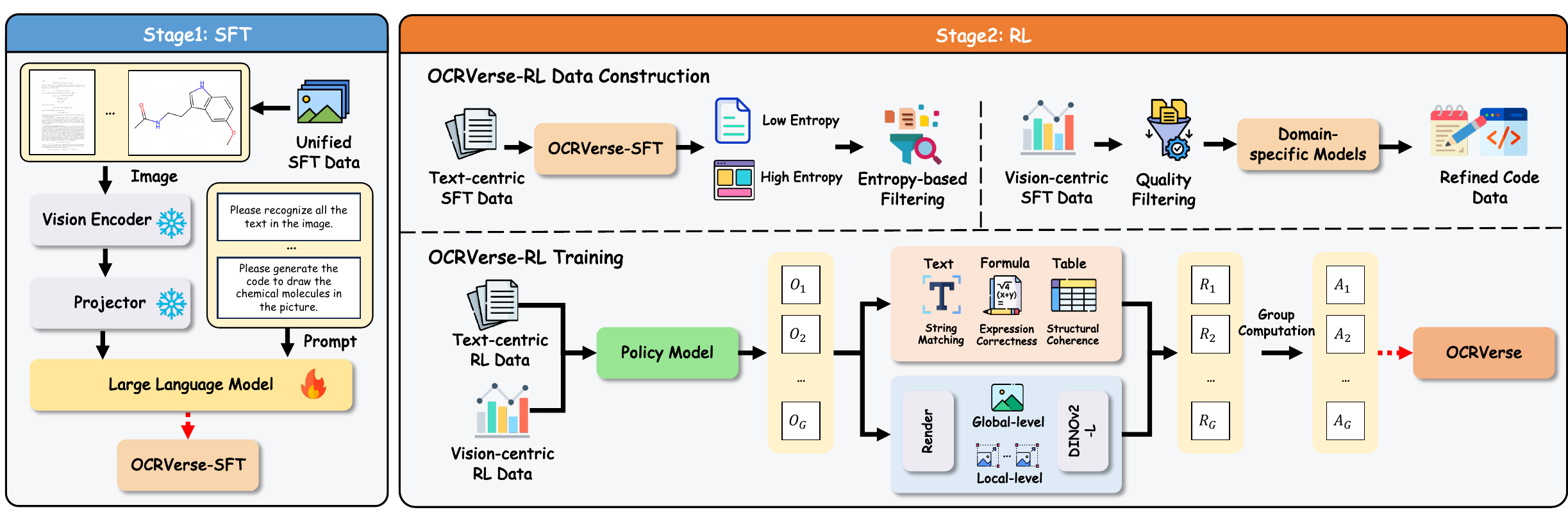}
\caption{OCRVerse training pipeline. \textbf{Stage 1:} SFT with unified cross-domain data. \textbf{Stage 2:} RL with domain-specific data construction and personalized reward mechanisms for text-centric (rule-based) and vision-centric (visual fidelity) optimization.}
\label{fig:model_pipeline}
\end{figure*}

In this section, we present the two-stage training methodology for OCRVerse. As shown in Figure~\ref{fig:model_pipeline}, the training process consists of SFT to establish foundational cross-domain knowledge and RL to optimize domain-specific performance through personalized reward mechanisms.

\subsection{SFT Stage}

The SFT stage aims to establish foundational cross-domain knowledge by directly mixing data from all eight domains, enabling the model to learn diverse visual patterns and output formats across both text-centric and vision-centric scenarios.

\subsubsection{Training Objective}

We fine-tune the pre-trained Qwen3-VL-4B~\cite{Qwen3-VL} model using standard autoregressive language modeling. The training objective is formulated as:

\begin{eqnarray}
\mathcal{L}_{\text{SFT}}(\theta) = -\mathbb{E}_{(x,y)\sim\mathcal{D}_{\text{SFT}}} \sum_{t=1}^{T} \log P_\theta(y_t | x, y_{<t})
\end{eqnarray}

where $x$ represents the input image, $y$ denotes the target output sequence, and $\theta$ are the model parameters. During training, we freeze the visual encoder and vision-language adapter while updating only the language model parameters, thereby preserving strong visual representations while focusing computational resources on improving text generation and format compliance.

\subsubsection{Cross-Domain Data Mixing}

To build a unified representation space, we directly mix data from all eight domains during SFT, including text-centric domains such as documents, tables, and formulas, as well as vision-centric domains such as charts, web pages, and scientific plots. This approach enables the model to learn shared visual-semantic patterns across heterogeneous data types while maintaining domain-specific output capabilities. 

\subsection{RL Stage}

While SFT establishes strong baseline performance, it faces limitations in handling domain-specific requirements and format-intensive content. The RL stage addresses these challenges by designing personalized reward strategies tailored to the characteristics of each domain, thereby resolving domain conflicts and optimizing specialized performance.

\subsubsection{Domain-Specific Reward Design}

To resolve domain-specific conflicts and optimize personalized performance, we design customized reward strategies tailored to the unique characteristics of each domain. This approach enables fine-grained optimization for different content types and output formats, thereby improving cross-domain fusion while avoiding conflicts from uniform reward signals.

\paragraph{Text-centric Domains.} For text-centric domains, we employ rule-based reward functions that separately evaluate different content types. For plain text, we use one minus normalized edit distance as the reward. For formulas, we first normalize the LaTeX expressions following OmniDocBench v1.5~\cite{ouyang2025omnidocbench} protocol and then compute BLEU score as the reward. For tables, we normalize the table structure according to OmniDocBench v1.5~\cite{ouyang2025omnidocbench} and apply TEDS-S metric as the reward. The overall text-centric reward is computed as:

\begin{eqnarray}
R_{\text{text}} = \frac{1}{|\mathcal{C}_{\text{valid}}|} \sum_{c \in \mathcal{C}} \mathbb{I}[|\text{GT}_c| > 0] \cdot r_c(\text{Pred}_c, \text{GT}_c)
\end{eqnarray}

where $\mathcal{C}$ represents the set of content types, $\mathcal{C}_{\text{valid}}$ denotes content types present in the ground truth, and $r_c(\cdot, \cdot)$ is the type-specific reward function.

\paragraph{Vision-centric Domains.} For vision-centric domains, we design visual fidelity rewards that measure perceptual similarity between rendered outputs and ground truth images. We leverage pre-trained DINOv2~\cite{oquab2023dinov2} encoder to extract visual features and compute cosine similarity in the embedding space. To handle images of varying resolutions, we employ a multi-scale reward mechanism that combines global-level similarity from downsampled thumbnails and local-level similarity from image patches:

\begin{eqnarray}
R_{\text{vision}} = \omega_{\text{global}} \cdot s_{\text{global}} + \omega_{\text{local}} \cdot \frac{1}{N} \sum_{i=1}^{N} s_{\text{local}}^{(i)}
\end{eqnarray}

where $s_{\text{global}}$ and $s_{\text{local}}^{(i)}$ measure global and local similarities respectively, and $\omega_{\text{global}}, \omega_{\text{local}}$ are weighting coefficients. Additionally, we introduce format alignment rewards to ensure generated code matches the expected programming language.

Through these domain-specific reward mechanisms, the model receives targeted feedback on format correctness, structural validity, and visual fidelity, enabling effective optimization for diverse domain requirements while maintaining coherent cross-domain performance.

\subsubsection{RL Data Construction}

To construct effective RL training data, we carefully select samples from the SFT dataset based on domain-specific criteria to ensure training effectiveness. For text-centric domains, we employ entropy-based filtering to select challenging samples that exhibit high structural complexity and prediction uncertainty, thereby focusing RL training on instances that require improved reasoning capabilities. For vision-centric domains, we collect diverse datasets spanning charts, web pages, SVG graphics, scientific plots, and chemical structures from multiple sources, applying quality filtering and refinement procedures to ensure high fidelity and visual complexity. This careful curation process results in a balanced RL dataset that covers both text-centric and vision-centric scenarios, providing sufficient diversity and difficulty to drive effective policy optimization through domain-specific reward signals.

\subsubsection{Policy Optimization}

We employ Group Relative Policy Optimization (GRPO)~\cite{shao2024deepseekmath} to fine-tune the model using domain-specific rewards. For each input $x$, we sample $G$ responses $\{o_1, o_2, \ldots, o_G\}$ from the current policy $\pi_{\theta_{\text{old}}}$ and compute their rewards $\{R_1, R_2, \ldots, R_G\}$. The group-normalized advantage for response $o_i$ is:

\begin{eqnarray}
A_i = \frac{R_i - \mu_G}{\sigma_G}
\end{eqnarray}

where $\mu_G$ and $\sigma_G$ are the mean and standard deviation of rewards within the group. The policy is optimized by maximizing:

\begin{eqnarray}
\mathcal{L}_{\text{RL}}(\theta) = \mathbb{E}_{x \sim \mathcal{D}_{\text{RL}}, \{o_i\}_{i=1}^{G} \sim \pi_{\theta_{\text{old}}}} \Bigg[ \frac{1}{G} \sum_{i=1}^{G} \min \Big( \rho_i A_i, \text{clip}(\rho_i, 1-\epsilon, 1+\epsilon) A_i \Big) \Bigg]
\end{eqnarray}

where $\rho_i = \frac{\pi_\theta(o_i|x)}{\pi_{\theta_{\text{old}}}(o_i|x)}$ is the probability ratio and $\epsilon$ is the clipping threshold. This optimization process enables the model to learn from domain-specific feedback while maintaining training stability through ratio clipping.

Through this two-stage training methodology, OCRVerse effectively establishes cross-domain knowledge during SFT and refines domain-specific capabilities during RL, achieving seamless fusion across diverse data types while avoiding conflicts that arise from naive multi-task learning.

\section{Experiment}

In this section, we present a comprehensive empirical evaluation of OCRVerse to demonstrate its efficacy across the diverse landscape of holistic OCR. We benchmark our model against a representative suite of baselines, including both frontier general-purpose multimodal models and specialized state-of-the-art OCR systems. Our evaluation is structured into two primary dimensions to reflect the unified capabilities of OCRVerse. First, Section~\ref{sec:text-centric eval} analyzes text-centric OCR performance, specifically focusing on document recognition accuracy across plain text, mathematical formulas, complex tables, and reading order recovery. Second, Section~\ref{sec:vision-centric eval} investigates vision-centric OCR capabilities, evaluating the model's proficiency in translating charts, web pages, SVG icons, geometric diagrams and molecular structures into structured code representations across multiple public benchmarks.

\subsection{Text-Centric Evaluation}
\label{sec:text-centric eval}
We evaluate OCRVerse's text-centric OCR on OmniDocBench v1.5~\cite{ouyang2025omnidocbench}, a comprehensive benchmark that assesses model performance across varied document types, quality levels, and parsing complexities to validate robustness and generalization in real-world applications.

\subsubsection{Experimental Settings}

\noindent \textbf{Evaluation benchmark.} We adopt OmniDocBench v1.5~\cite{ouyang2025omnidocbench} as the primary testbed to comprehensively evaluate the document parsing capabilities of OCRVerse. This benchmark serves as a rigorous standard to validate model robustness and generalization across real-world applications. OmniDocBench v1.5 is an expanded dataset comprising 1,355 document pages, enriched with 374 additional pages compared to the previous version. It features a balanced distribution of bilingual content in both Chinese and English and covers nine diverse document types—including academic papers, textbooks, financial reports, and exam papers. By incorporating varied layout structures (from single- to multi-column) and rich elements such as text, mathematical formulas, and structured tables, the benchmark enables a thorough assessment of OCRVerse in handling complex parsing scenarios.

\noindent \textbf{Metrics.} Adhering to the official evaluation protocol of OmniDocBench v1.5~\cite{ouyang2025omnidocbench}, we employ three specialized metrics to comprehensively assess different dimensions of document parsing: Edit Distance~\cite{lcvenshtcin1966binary} for text recognition, Character Detection Matching (CDM)~\cite{wang2024cdm} for mathematical formula recognition, and Tree Edit Distance-based Similarity (TEDS)~\cite{zhong2020image} for table recognition. To provide a holistic performance assessment, we report the Overall Score, which aggregates these metrics as follows:

\begin{equation}
\text{Overall} = \frac{(1 - \text{Text}^{\text{Edit}}) \times 100 + \text{Table}^{\text{TEDS}} + \text{Formula}^{\text{CDM}}}{3}.
\end{equation}

\noindent \textbf{Comparison.}
We conduct a comprehensive comparative analysis of our proposed method against a wide spectrum of state-of-the-art OCR approaches. These methods are categorized into three distinct groups: 
(a) Pipeline Tools, such as Marker~\cite{paruchuri2025marker}, Mineru2-pipeline~\cite{wang2024mineru}, and PP-StructureV3~\cite{cui2025paddleocr3}, which utilize separate modules for layout analysis and text recognition. (b) General VLMs, including large-scale multimodal models like GPT-4o~\cite{achiam2023gpt4}, InternVL3~\cite{wang2025internvl35}, and Qwen2.5-VL~\cite{bai2025qwen25vl}. (c) Specialized VLMs, which are specifically optimized for document parsing tasks, represented by models such as Deepseek-OCR~\cite{wei2025deepseekocr}, dots.ocr~\cite{li2025dots}, and PaddleOCR-VL~\cite{cui2025paddleocrvl}.

\subsubsection{Quantitative Results}

\begin{table*}[!ht]
\centering
\caption{Performance comparison on OmniDocBench v1.5~\cite{ouyang2025omnidocbench}. \textbf{Bold} and \underline{underline} denote the best and second-best performance among end-to-end specialized VLMs, respectively. RD: Release Date, E2E: End-to-end (\ding{51}) or pipeline-based (\ding{55}), RO: Reading Order.}
\label{tab:omnidocbench}
\resizebox{\textwidth}{!}{
\begin{tabular}{llccccccccc}
\toprule
\textbf{Model Type} & \textbf{Methods} & \textbf{RD} & \textbf{E2E} & \textbf{Overall}$\uparrow$ & \textbf{Text}$^{\text{Edit}}\downarrow$ & \textbf{Formula}$^{\text{CDM}}\uparrow$ & \textbf{Table}$^{\text{TEDS}}\uparrow$ & \textbf{Table}$^{\text{TEDS-S}}\uparrow$ & \textbf{RO}$^{\text{Edit}}\downarrow$ \\
\midrule
\multirow{3}{*}{Pipeline Tools} 
& Marker-1.8.2~\cite{paruchuri2025marker} & 2025 & \ding{55} & 71.30 & 0.206 & 76.66 & 57.88 & 71.17 & 0.250 \\
& Mineru2-pipeline~\cite{wang2024mineru} & 2025 & \ding{55}  & 75.51 & 0.209 & 76.55 & 70.90 & 79.11 & 0.225 \\
& PP-StructureV3~\cite{cui2025paddleocr} & 2025 & \ding{55}  & 86.73 & 0.073 & 85.79 & 81.68 & 89.48 & 0.073 \\
\midrule
\multirow{5}{*}{General VLMs} 
& GPT-4o~\cite{achiam2023gpt} & 2023 & \ding{51} & 75.02 & 0.217 & 79.70 & 67.07 & 76.09 & 0.148 \\
& InternVL3-76B~\cite{zhu2025internvl3} & 2025 & \ding{51}  & 80.33 & 0.131 & 83.42 & 70.64 & 77.74 & 0.113 \\
& InternVL3.5-241B~\cite{wang2025internvl3} & 2025 & \ding{51}  & 82.67 & 0.142 & 87.23 & 75.00 & 81.28 & 0.125 \\
& Qwen2.5-VL-72B~\cite{Qwen2.5-VL} & 2025 & \ding{51}  & 87.02 & 0.094 & 88.27 & 82.15 & 86.22 & 0.102 \\
& Gemini-2.5 Pro~\cite{comanici2025gemini} & 2025  & \ding{51}  & 88.03 & 0.075 & 85.82 & 85.71 & 90.29 & 0.097 \\
\midrule
\multirow{15}{*}{Specialized VLMs} 
& Dolphin~\cite{feng2025dolphin} & 2025.05 & \ding{55} & 74.67 & 0.125 & 67.85 & 68.70 & 77.77 & 0.124 \\
& MinerU2-VLM~\cite{wang2024mineru} & 2025.06 & \ding{55} & 85.56 & 0.078 & 80.95 & 83.54 & 87.66 & 0.086 \\
& MonkeyOCR-pro-1.2B~\cite{li2025monkeyocr} & 2025.07 & \ding{55} & 86.96 & 0.084 & 85.02 & 84.24 & 89.02 & 0.130 \\
& MonkeyOCR-3B~\cite{li2025monkeyocr} & 2025.06 & \ding{55} & 87.13 & 0.075 & 87.45 & 81.39 & 85.92 & 0.129 \\
& MonkeyOCR-pro-3B~\cite{li2025monkeyocr} & 2025.07 & \ding{55} & 88.85 & 0.075 & 87.25 & 86.78 & 90.63 & 0.128 \\
& MinerU2.5~\cite{niu2025mineru2} & 2025.09 & \ding{55} & 90.67 & 0.047 & 88.46 & 88.22 & 92.38 & 0.044 \\
& PaddleOCR-VL~\cite{cui2025paddleocr-vl} & 2025.10 & \ding{55} & 92.56 & 0.035 & 91.43 & 89.76 & 93.52 & 0.043 \\
\cmidrule{2-10}
& OCRFlux-3B~\cite{ocrflux2025} & 2025.06 & \ding{51} & 74.82 & 0.193 & 68.03 & 75.75 & 80.23 & 0.202 \\
& Mistral OCR~\cite{mistral2025ocr} & 2025.03 & \ding{51} & 78.83 & 0.164 & 82.84 & 70.03 & 78.04 & 0.144 \\
& POINTS-Reader~\cite{liu2025points} & 2025.08 & \ding{51} & 80.98 & 0.134 & 79.20 & 77.13 & 81.66 & 0.145 \\
& olmOCR-7B~\cite{poznanski2025olmocr-1} & 2025.02 & \ding{51} & 81.79 & 0.096 & {86.04} & 68.92 & 74.77 & 0.121 \\
& Nanonets-OCR-s~\cite{mandalm2025nanonets} & 2025.06 & \ding{51} & 85.59 & 0.093 & 85.90 & 80.14 & 85.57 & 0.108 \\
& Deepseek-OCR~\cite{wei2025deepseek} & 2025.10 & \ding{51} & 87.01 & 0.073 & 83.37 & {84.97} & {88.80} & 0.086 \\
& dots.ocr~\cite{rednote2025dotsocr} & 2025.07 & \ding{51} & {88.41} & \underline{0.048} & 83.22 & {86.78} & 90.62 & \textbf{0.053} \\

& FD-RL~\cite{zhong2025reading} & 2025.11 & \ding{51} & {90.41} & {0.049} & {88.67} & {87.35} & \textbf{92.10} & \underline{0.055} \\

& HunyuanOCR~\cite{hunyuanvisionteam2025hunyuanocrtechnicalreport} & 2025.11 & \ding{51} & \textbf{94.10} & \textbf{0.042} & \textbf{94.73} & \textbf{91.81} & \textbf{-} & {-} \\

& Deepseek-OCR2~\cite{wei2026deepseekocr} & 2026.01 & \ding{51} & \underline{91.09} &  0.048 &  \underline{90.31} & \underline{87.75} & \underline{92.06} & 0.057 \\

& OCRVerse (Ours) & 2026.01 & \ding{51} & {89.23} & {0.052} & {87.13} & {85.77} & {90.35} & 0.068 \\

\bottomrule
\end{tabular}
}
\end{table*}

As shown in Table~\ref{tab:omnidocbench}, OCRVerse achieves an overall score of 89.23 on OmniDocBench v1.5, ranking among the top-tier end-to-end specialized VLMs. Notably, OCRVerse substantially surpasses general-purpose VLMs including Gemini-2.5 Pro (88.03) and Qwen2.5-VL-72B (87.02), despite having significantly fewer parameters. This performance validates the efficacy of our holistic training paradigm in bridging the gap between general visual understanding and specialized document parsing.

In formula recognition, OCRVerse achieves a CDM score of 87.13, outperforming several larger end-to-end models such as Deepseek-OCR (83.37) and olmOCR-7B (86.04). This superior performance validates the effectiveness of our systematic synthetic formula data strategy, which incorporates single-line formulas for basic expressions, multi-line formulas for complex derivations, and page-level formulas for comprehensive mathematical content. Furthermore, our training data covers diverse academic disciplines including mathematics, physics, and computer science, with difficulty levels ranging from undergraduate to graduate. Such comprehensive coverage enables robust recognition of complex mathematical expressions across real-world scenarios.

For text recognition and reading order, OCRVerse achieves competitive results with an edit distance of {0.052} for text and {0.068} for reading order. However, compared to layout-aware models like dots.ocr (0.048 for text, 0.053 for reading order), there remains a performance gap. Currently, OCRVerse does not incorporate explicit layout-aware mechanisms, which limits its ability to capture fine-grained spatial relationships in complex document structures. In future work, we plan to explore the use of region-level OCR data more deeply to enhance the model's layout-aware capabilities, which is expected to further improve performance in text recognition and reading order preservation across diverse document layouts.

In table recognition tasks, OCRVerse achieves a TEDS score of 85.77 and TEDS-S score of 90.35, which lags behind Deepseek-OCR2 and HunyuanOCR. This performance gap suggests opportunities for further enhancement through more comprehensive table data coverage. In future work, we plan to increase our focus on table data construction, particularly by supplementing training samples with complex table structures involving multi-row, multi-column layouts and cross-row, cross-column cell spanning. Such enriched table data is expected to significantly improve the model's capability in handling intricate table configurations.

These results collectively demonstrate that OCRVerse achieves strong empirical performance across diverse document parsing tasks with remarkable parameter efficiency. The competitive results across formula, text, and table recognition—combined with the successful unification of both text-centric and vision-centric capabilities within a single lightweight architecture—suggest the substantial potential of our holistic OCR paradigm as a scalable foundation for next-generation document intelligence systems.

\subsection{Vision-Centric Evaluation}
We conduct a comprehensive evaluation of OCRVerse on vision-centric OCR tasks, where the objective transcends mere text recognition to encompass the interpretation of visually information-dense images and their translation into executable code or structured representations. This assessment spans five diverse benchmarks, encompassing chart-to-code generation, web layout reconstruction, scalable vector graphics synthesis, mathematical formula recognition, and chemical structure parsing.

\label{sec:vision-centric eval}
\subsubsection{Experimental Settings}

\noindent \textbf{Benchmarks.} To comprehensively evaluate the vision-centric OCR capabilities of our proposed OCRVerse, we conducted extensive experiments across five diverse structured Image-to-Code benchmarks. These benchmarks cover a wide spectrum of visual domains, assessing the model's ability to translate visually dense information into executable code or structured representations. Specifically, the evaluation tasks include: (1) ChartMimic~\cite{yang2024chartmimic} for direct chart-to-code generation; (2) Design2Code~\cite{si2025design2code}, which evaluates the precision of reproducing web layouts in the web-to-HTML task; (3) UniSVG~\cite{li2025unisvg}, assessing the generation of scalable vector graphics in the image-to-SVG task; (4) Image2Struct~\cite{roberts2024image2struct}, testing the conversion of scientific documents and formulas in the image-to-LaTeX task; (5) ChemDraw~\cite{zhao2025vincicoder}, focusing on the recognition of chemical structures in the molecule-to-code task.

\noindent \textbf{Metrics.} For all benchmarks, we strictly adhere to the evaluation protocols officially defined in their respective GitHub repositories or original papers. For ChartMimic~\cite{yang2024chartmimic}, we report the code execution success rate and the average of low-level metrics (Text, Layout, Type, and Color Scores). For high-level evaluation, we employ the GPT-4o Score. Regarding UniSVG~\cite{li2025unisvg}, we present the low-level score, computed as the average of SSIM and (1 - LPIPS), alongside the high-level CLIP similarity. For Design2Code~\cite{si2025design2code}, we report both the CLIP similarity (high-level) and the element-matching scores (low-level) proposed by the benchmark authors. For Image2Struct~\cite{roberts2024image2struct}, we evaluate using the earth mover's similarity (EMS) and the rendering success rate. Finally, for ChemDraw~\cite{zhao2025vincicoder}, we report the code execution success rate and the Tanimoto similarity.

\noindent \textbf{Comparison.} 
To demonstrate the competitive performance of OCRVerse in vision-centric OCR tasks, we benchmark our method against a wide array of state-of-the-art baselines. These models are categorized into two groups: (1) Closed-Source Models, including leading proprietary large multimodal models such as Gemini-2.5-Pro~\cite{comanici2025gemini}, Claude-4.5-Sonnet~\cite{anthropic2025claude}, and GPT-5~\cite{openai2025gpt5}; and (2) Open-Source Models, covering recent high-performing series such as InternVL3~\cite{zhu2025internvl3} (\textit{e.g.}, 8B, 14B, 38B) and Qwen-VL (\textit{e.g.}, Qwen2.5-VL~\cite{bai2025qwen25vl}, Qwen3-VL~\cite{Qwen3-VL}).

\subsubsection{Quantitative Results}
Table~\ref{tab:main_results} presents the comprehensive evaluation results of OCRVerse against state-of-the-art baselines across five vision-centric OCR benchmarks. Our 4B-parameter model demonstrates remarkably competitive performance, even surpassing significantly larger counterparts in several key metrics.

On ChartMimic, OCRVerse achieves 84.8\% execution success rate, substantially outperforming open-source models of comparable size (\textit{e.g.}, Qwen3-VL-8B: 78.3\%, InternVL3-8B: 63.3\%). Notably, our low-level score (72.2) and high-level score (75.4) exceed those of Qwen2.5-VL-72B (72.7 and 79.1 respectively) despite being 18$\times$ smaller, suggesting superior fine-grained visual understanding. 

For UniSVG, OCRVerse attains a composite score of 76.3, ranking second only to GPT-5 (77.3) among all evaluated models, with particularly strong high-level CLIP similarity (85.2 vs. 88.3 for GPT-5). This indicates our model's exceptional capability in preserving semantic consistency between visual inputs and generated SVG primitives.

Regarding Design2Code, OCRVerse achieves competitive low-level (85.7) and high-level (87.4) scores, validating its proficiency in web layout reconstruction. The most striking results emerge on Image2LaTeX-plot, where OCRVerse significantly outperforms all baselines with 88.7\% rendering success rate and 63.1 EMS—surpassing GPT-5 (78.7\%, 57.4) by large margins. This dominance in scientific plot interpretation highlights the efficacy of our SFT-RL training strategy in capturing complex hierarchical structures. Similarly, on ChemDraw, OCRVerse achieves 89.1\% execution success rate and 54.7 Tanimoto similarity, outperforming all open-source alternatives and approaching GPT-5 in molecular structure recognition.

These results collectively demonstrate that OCRVerse achieves superior parameter efficiency, delivering performance comparable to or exceeding 70B parameter models with merely 4B parameters. This suggests the substantial potential of our holistic OCR paradigm and multi-domain training methodology in developing lightweight yet powerful vision-language models for diverse code generation tasks.

\begin{table*}[t]
\setlength{\tabcolsep}{3pt}
\caption{Evaluation results of comparing OCRVerse with various baseline models on multimodal code generation benchmarks.}

\label{tab:main_results}
\centering
\resizebox{\textwidth}{!}{
\begin{tabular}{lc|ccc|ccc|cc|cc|cc}
\toprule
\multirow{2}{*}{Model}  & \multirow{2}{*}{Parameters} & \multicolumn{3}{c|}{\textbf{ChartMimic}} & \multicolumn{3}{c|}{\textbf{UniSVG-ISVGEN}} & \multicolumn{2}{c|}{\textbf{Design2Code}} & 
\multicolumn{2}{c|}{\textbf{Image2Latex-plot}} & \multicolumn{2}{c}{\textbf{ChemDraw}} \\
\cmidrule{3-14}
   & & Exec.Rate & Low-L  & High-L & Low-L  & High-L  & Score & Low-L & High-L & Ren.Succ. & EMS & Exec.Rate & Tani.Sim.\\ 
\midrule

\multicolumn{13}{l}{\textbf{Closed-Source Models}} \\ 
\midrule
Gemini-2.5-Pro & - &97.3 & 88.7 & 83.8 & 53.6 & 80.3 & 69.6  & 90.8 & 91.4 & 74.3 & 52.5 &77.3& 2.8\\
Claude-4.5-Sonnet & - &97.8 & 89.6 & 82.9& 61.0 & 83.4 & 74.6 & 90.4 & 90.8& 72.7 & 50.2 & 95.3 & 41.7\\
GPT-5 & - &94.8 & 81.9 &  78.3 & 60.8 & 88.3 & 77.3 & 90.6 & 91.0 & 78.7 & 57.4 & 93.8 & 52.1\\
\midrule
\multicolumn{13}{l}{\textbf{Open-Source Models}} \\
\midrule
Qwen2.5-VL-7B  & 7B & 68.7 & 42.2 & 40.1 &  47.5 & 73.8 & 63.3 & 83.4 & 87.6 & 42.7 & 25.5 &21.1 & 11.7 \\
InternVL3-8B & 8B & 63.3 & 43.8 & 46.1 & 54.5 & 77.4 & 68.2 & 85.3 & 87.6 & 57.7& 38.6 & 42.2 & 6.2 \\
Qwen3-VL-8B & 8B & 78.3 & 62.5 & 67.8 & 53.0 & 77.0 & 67.4 & 85.5  & 87.2 & 47.7 & 33.0 & 78.9 & 41.2 \\
InternVL3.5-8B & 8B &  66.7 & 46.0 & 48.3 & 55.0 & 78.0 & 68.6 & 85.8 & 87.3 & 58.3 & 40.5 &49.2 & 7.8\\
InternVL3-14B  & 14B &  72.3 & 51.3 & 54.1 & 51.4 & 75.5 & 65.8 & 85.8 & 87.5 & 73.3 & 52.2 & 71.1  & 40.2 \\
InternVL3.5-14B & 14B & 73.2 & 52.8 & 55.4 & 52.0 & 75.0 & 65.9 & 86.1 & 87.8 & 73.0 & 50.2  & 71.9 & 39.3 \\
Qwen3-VL-32B  & 32B & 83.0 & 66.9 & 77.5 & 68.0 & 86.0 & 78.8 & 88.6 & 89.8 & 75.7 & 53.3 & 37.5 & 48.8\\
InternVL3.5-38B & 38B & 79.0 &60.0 & 71.8 & 51.9 & 77.3 & 67.1& 87.8 & 88.4 & 72.6 & 49.5 &55.5 & 31.4\\
Qwen2.5-VL-72B & 72B & 88.5 & 72.7 & 79.1 & 47.7 & 76.0 & 64.7 & 86.9 & 88.7 & 62.0 & 41.7 & 75.8 & 28.0\\
\midrule
OCRVerse (Ours) & 4B & 84.8 & 72.2	& 75.4 &  63.2 & 85.2 & 76.3 & 85.7 &87.4 & 88.7 & 63.1 & 89.1 & 54.7\\
\bottomrule 
\end{tabular}
}
\end{table*}

\section{Conclusion}
In this technical report, we propose OCRVerse, the first holistic OCR method that unifies text-centric and vision-centric capabilities in an end-to-end manner. Through comprehensive data engineering across diverse domains and an innovative two-stage SFT-RL training methodology, OCRVerse bridges character-level recognition with code-level representation while resolving cross-domain conflicts through personalized reward strategies. Experimental results demonstrate competitive performance across both text-centric and vision-centric scenarios, achieving 89.23 on OmniDocBench v1.5 while matching open-source models on vision-centric benchmarks. By advancing OCR technology from fragmented approaches to holistic recognition, OCRVerse provides a practical solution for real-world applications including data visualization, web page analysis, and intelligent content understanding. We release the model to promote further research and facilitate broader adoption of holistic OCR capabilities in multimodal AI systems.

\newpage
\bibliographystyle{unsrt}
\bibliography{main}

\newpage











\end{document}